\documentclass[letterpaper, 10 pt, conference]{ieeeconf}
\IEEEoverridecommandlockouts
\usepackage{multicol}
\usepackage{graphicx}
\usepackage{xspace}
\usepackage{xcolor}
\usepackage{caption}

\usepackage{pifont}

\usepackage{wrapfig}
\usepackage{bbding}  % For \Checkmark
\usepackage{pifont}  % For \XSolidBrush
\usepackage[bookmarks=true]{hyperref}

\usepackage{booktabs}
\usepackage{float}
\usepackage{amsmath}  % 提供了高级数学公式支持
\usepackage{amssymb}  % 提供了额外的数学符号
\usepackage{multirow, booktabs}
\usepackage{siunitx}

\newcommand{\ours}[0]{iDP3\xspace}
\newcommand{\website}[0]{\href{https://humanoid-manipulation.github.io
}{our website}\xspace}

\usepackage{colortbl}
\definecolor{ourcolor}{HTML}{99e0eb}
\definecolor{ourblue}{HTML}{27a2c3}

\definecolor{tablecolor}{HTML}{ccf2f5} 

\definecolor{tablecolor2}{HTML}{ffcdb4}
\definecolor{citecolor}{HTML}{fe7b5b}
\definecolor{grey}{rgb}{0.9, 0.9, 0.9}
\usepackage{amssymb}

\usepackage{listings}
\definecolor{gred}{rgb}{0.859,0.267,0.216}
\definecolor{ggreen}{rgb}{0.059,0.616,0.345}

\definecolor{deepblue}{HTML}{27a2c3}

\definecolor{deepred}{HTML}{7c2320}

\usepackage{cite}
\usepackage{amsmath,amssymb,amsfonts}
\usepackage{algorithmic}
\usepackage{graphicx}
\usepackage{textcomp}
\usepackage{xcolor}

\definecolor{deepgreen}{RGB}{63, 126, 49}
\definecolor{deepred2}{RGB}{196, 49, 25}

\newcommand{\cmark}{\textcolor{deepgreen}{\ding{51}}}%
\newcommand{\xmark}{\textcolor{deepred2}{\ding{55}}}%

\begin{document}

\title{\LARGE{\textbf{
Generalizable Humanoid Manipulation with 3D Diffusion Policies
}}}

\author{
Yanjie Ze$^{1}$\quad Zixuan Chen$^{2}$\quad Wenhao Wang$^{3}$\quad Tianyi Chen$^{3}$\vspace{0.03in}\\ Xialin He$^{4}$\quad Ying Yuan$^{5}$\quad
Xue Bin Peng$^2$\quad Jiajun Wu$^1$\vspace{0.03in}\\
$^1$Stanford University\quad $^2$Simon Fraser University \quad
$^3$UPenn\quad
$^4$UIUC\quad $^{5}$CMU\vspace{0.05in}\\
\href{https://humanoid-manipulation.github.io}{\color{deepred}\textbf{\textsc{humanoid-manipulation.github.io}}\xspace}
}

\twocolumn[{%
\renewcommand\twocolumn[1][]{#1}%
\maketitle
\vspace{-0.2in}
\begin{center}
    \centering
    \captionsetup{type=figure}
    \includegraphics[width=1.0\textwidth]{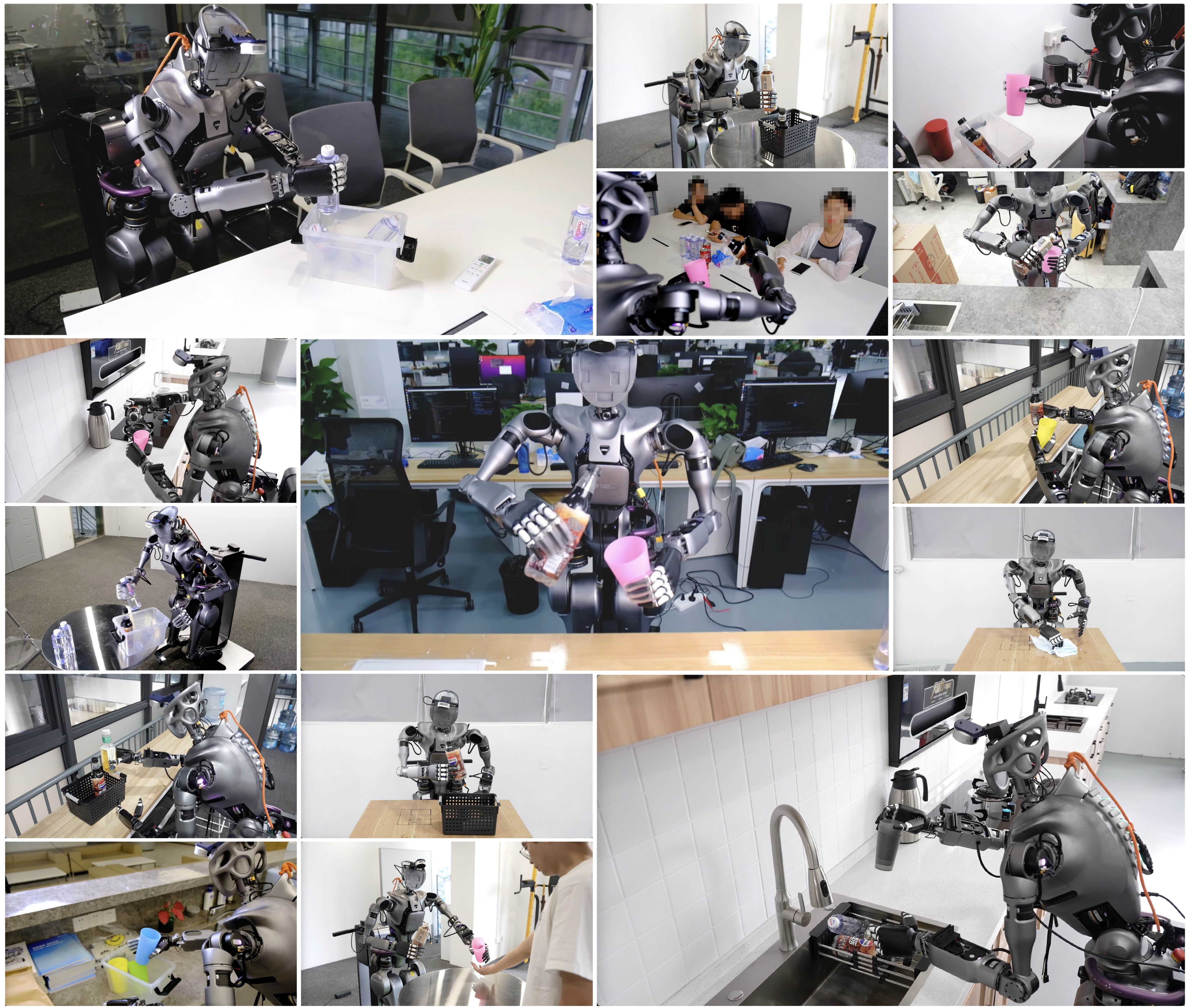}
     \vspace{-0.2in}
    \caption{\textbf{Humanoid manipulation in diverse unseen scenarios.} With our system, we are able to 1) collect human-like imitation learning data and 2) enable a full-sized humanoid robot to perform useful skills in \textit{diverse} real-world environments using data only from a \textit{single} scene,. \textit{The scenes are not cherry-picked.} Videos are available on \website.}
    \label{fig:diverse scene}
\end{center}
\vspace{-0.05in}
}]

\begin{abstract}
Humanoid robots capable of autonomous operation in diverse environments have long been a goal for roboticists. However, autonomous manipulation by humanoid robots has largely been restricted to one specific scene, primarily due to the difficulty of acquiring generalizable skills and the expensiveness of in-the-wild humanoid robot data. In this work, we build a real-world robotic system to address this challenging problem. Our system is mainly an integration of 1) a whole-upper-body robotic teleoperation system to acquire human-like robot data, 2) a 25-DoF humanoid robot platform with a height-adjustable cart and a 3D LiDAR sensor, and 3) an improved 3D Diffusion Policy learning algorithm for humanoid robots to learn from noisy human data. We run more than 2000 episodes of policy rollouts on the real robot for rigorous policy evaluation. Empowered by this system, we show that using only data collected in one single scene and with only onboard computing, a full-sized humanoid robot can autonomously perform skills in diverse real-world scenarios. Videos are available at \href{https://humanoid-manipulation.github.io}{humanoid-manipulation.github.io}.

% Recent advances in 3D visuomotor policies, such as the 3D Diffusion Policy (DP3), have shown promise in extending these capabilities to wilder environments. However, 3D visuomotor policies often rely on camera calibration and point-cloud segmentation, which present challenges for deployment on mobile robots like humanoids. In this work, we introduce the Improved 3D Diffusion Policy (iDP3), a novel 3D visuomotor policy that eliminates these constraints by leveraging egocentric 3D visual representations. 

% We demonstrate that iDP3 enables a full-sized humanoid robot to autonomously perform skills in diverse real-world scenarios, using only data collected in the lab.

\end{abstract}

\section{Introduction}

\begin{table*}[ht]
\centering
\caption{Compared to recent real-world robot learning systems for humanoid robots and dexterous manipulation, our work focuses on developing a humanoid learning system that generalizes the learned policy to unseen real-world scenes—an aspect that has been missing in previous humanoid works.}
\label{tab:teleop_comparison}
\resizebox{1.0\textwidth}{!}{%
\rowcolors{6}{gray!15}{white}
\begin{tabular}{lccccccccccc}
\toprule
 & \multicolumn{4}{c}{\textbf{Teleoperation}} 
 & \multicolumn{3}{c}{\textbf{ Generalization Abilities}}
 & \multicolumn{1}{c}{\textbf{Rigorous Policy Evaluation}} \\
\cmidrule(lr){2-5} \cmidrule(lr){6-8} \cmidrule(lr){9-9}
\textbf{Method} 
 & \textbf{Arm\&Hand} 
 & \textbf{Head} 
 & \textbf{Waist} & \textbf{Leg}
 & \textbf{Object} 
 & \textbf{Camera View} 
 & \textbf{Scene} 
 & \textbf{Real-World Episodes} \\
\midrule
AnyTeleop~\cite{qin2023anyteleop}  & \cmark & \xmark & \xmark & \xmark
  & \cmark & \xmark
  & \xmark
  & $0$ \\
  
DP3~\cite{Ze2024DP3} & \cmark & \xmark & \xmark & \xmark & \cmark &   \cmark & \xmark  & $186$ \\

  BunnyVisionPro~\cite{ding2024bunnyvisionpro}& \cmark & \xmark & \xmark & \xmark
  & \cmark& \xmark
  & \xmark
  & $540$\\
  
ACE~\cite{yang2024ace} 
  & \cmark & \xmark & \xmark & \xmark
  & \xmark& \xmark
  & \xmark
   & $60$ \\
  
  Bi-Dex~\cite{shaw2024bimanual}
   & \cmark & \xmark & \xmark & \xmark
  & \xmark & \xmark
  & \xmark
   & $50$\\

  OmniH2O~\cite{he2024omnih2o} & \cmark & \xmark & \xmark & \cmark & \xmark & \xmark & \xmark & $90$ \\
  
  HumanPlus~\cite{fu2024humanplus} & \cmark & \xmark & \xmark &  \cmark & \xmark & \xmark & \xmark   & $160$\\
  
  Hato~\cite{lin2024twisting} & \cmark & \xmark & \xmark & \xmark &  \xmark & \xmark & \xmark & $300$ \\
  ManiWhere~\cite{yuan2024maniwhere} & \cmark & \xmark & \xmark & \xmark  & \cmark & \cmark & \cmark & $200$\\
OpenTeleVision~\cite{cheng2024opentv} 
  & \cmark & \cmark & \xmark & \xmark
  & \xmark & \xmark
  & \xmark
  & $75$ \\
  
\textbf{This Work} 
  & \cmark & \cmark & \cmark & \xmark
  & \cmark & \cmark
  & \cmark
& $2253$ \\
\bottomrule
\end{tabular}%
}
\end{table*}

Robots capable of performing diverse tasks in unstructured environments have long been a significant goal in the robotics community, with the development of intelligent humanoid robots representing one promising pathway. Recently, substantial progress has been made in developing humanoid robot hardware~\cite{BD_Atlas, Tesla_Optimus, Figure01, Unitree_H1, Fourier_GR1} as well as teleoperation and learning systems for these robots~\cite{cheng2024opentv, yang2024ace, fu2024humanplus, he2024h2o, he2024omnih2o}. However, due to the limited generalization capabilities of the employed learning methods~\cite{chi2023diffusion_policy, zhao2023aloha, fu2024mobile_aloha, brohan2022rt1, brohan2023rt2} and the high cost of acquiring humanoid robot data from diverse scenes, these autonomous humanoid manipulation skills are all confined to their training scenarios and hard to generalize to new scenes~\cite{he2024omnih2o, he2024h2o, fu2024humanplus, cheng2024opentv, ding2024bunnyvisionpro, yang2024ace,shaw2024bimanual,lin2024hato}, as shown in Table~\ref{tab:teleop_comparison}.

% Recent advances in 3D visuomotor policies have shown great potential to generalize the learned skills to more complex and diverse scenarios~\cite{Ze2024DP3, 3d_diffuser_actor, grotz2024peract2, shridhar2023peract, Ze2023GNFactor}. Among these, the 3D Diffusion Policy (DP3, \cite{Ze2024DP3}) is effective in a variety of simulated and real-world tasks across different embodiments. These include deformable object manipulation with a dexterous hand~\cite{Ze2024DP3} or a mobile arm~\cite{yang2024equibot}, long-horizon bi-manual manipulation~\cite{ding2024bunnyvisionpro}, and loco-manipulation with a quadrupedal robot~\cite{he2024learning}. Despite DP3's generalizability, its applications have been restricted to tasks performed using a third-person view with a calibrated fixed camera, largely due to the need for accurate camera calibration and point-cloud segmentation, both of which are inherent challenges in 3D visuomotor policies.

In this work, we aim to develop \textit{a real-world humanoid robot learning system that can learn generalizable humanoid manipulation skills by 3D visuomotor policies}. An overview of our system is in Figure~\ref{fig:system}.

First, we design a humanoid robot learning platform, where a 29-DoF full-sized humanoid robot is fixed on a moveable and height-adjustable cart. This platform can stabilize humanoid robots even when the waist is leaning forward, so that we can safely utilize the waist DoF of humanoid robots. Besides, the robot head is attached with a 3D LiDAR sensor for generalizable policy learning.

Second, for human-like data collection, we design a whole-upper-body teleoperation system that maps human joints to a full-sized humanoid robot. Unlike the common bi-manual manipulation system, our teleoperation incorporates waist degrees of freedom and active vision, greatly expanding the robot's operational workspace, particularly when handling tasks at varying heights. We also stream real-time vision from LiDAR sensors to humans for egocentric teleoperation.

Third, to learn generalizable manipulation skills with egocentric human data, we re-formulate the third-person 3D learning algorithm 3D Diffusion Policy (DP3, \cite{Ze2024DP3}) to an egocentric version, eliminating the need for camera calibration and point cloud segmentation. By more than \textit{2000} real-world evaluation trials, we bring solid improvements over the original DP3 towards real-world humanoid manipulation. The resulting policy is termed as the Improved 3D Diffusion Policy (iDP3). Though this work only applies iDP3 on the Fourier GR1~\cite{Fourier_GR1} humanoid robot, we emphasize that iDP3 is a general 3D learning algorithm that can be applied to different robot platforms including mobile robots and humanoid robots.

Finally, we deploy our system to unseen real-world scenarios. We surprisingly found that, due to the robustness of our 3D representations and the flexibility of our platform, our policy \textit{zero-shot} generalize to a lot of randomly selected unseen scenarios, such as kitchens, meeting rooms, and offices, as shown in Figure~\ref{fig:diverse scene}.

% Through extensive real-world experiments and ablation studies, we demonstrate that iDP3 exhibits remarkable generalization across diverse scenes and shows strong view invariance, along with high effectiveness.

To summarize our contributions, we build a real-world humanoid robot system that can learn generalizable manipulation skills from only one single scene, utilizing 3D visuomotor policies. As far as we know, we are the first to successfully enable a full-sized humanoid robot to performs skills autonomously in diverse unseen scenes with data only from a single scene using 3D imitation learning.

% Our core contributions are summarized as follows: 
% \begin{itemize}%[leftmargin=*]
% \item We introduce the Improved 3D Diffusion Policy (iDP3), a 3D visuomotor policy that can be applied to any robot, supporting both egocentric and third-person views, while achieving high efficiency and strong generalization abilities. \item We develop a whole-upper-body teleoperation system for a humanoid robot, enabling efficient data collection from humans.
% \item We demonstrate that our policy deployed on a humanoid robot can successfully generalize contact-rich manipulation skills to a wide range of real-world scenarios, with data collected in a single scene.
% \end{itemize}

\section{Related Work}

The autonomous execution of diverse skills by humanoid robots in complex, real-world environments has long been a central goal in robotics. Recently, learning-based methods have shown promising progress toward this objective, particularly in the areas of locomotion~\cite{gu2024advancing, radosavovic2024humanoid_locomotion, radosavovic2024humanoid_locomotion_rl, zhuang2024humanoid_parkour, cheng2024expressive}, manipulation~\cite{cheng2024opentv, yang2024ace, wang2024dexcap}, and loco-manipulation~\cite{fu2024humanplus, seo2023trill, he2024h2o, he2024omnih2o}. While several works have successfully demonstrated humanoid locomotion in unstructured, real-world environments~\cite{gu2024advancing, radosavovic2024humanoid_locomotion, zhuang2024humanoid_parkour}, manipulation skills in unseen environments remain largely unexplored~\cite{cheng2024opentv, fu2024humanplus, he2024omnih2o}.

In Table~\ref{tab:teleop_comparison}, we list recent works that build real-world robotic systems for humanoid robots/dexterous manipulation. We found that existing works in humanoid robots~\cite{he2024omnih2o,ding2024bunnyvisionpro,yang2024ace,fu2024humanplus,lin2024hato,cheng2024opentv} miss the study of generalization abilities for humanoid manipulation, mainly due to the limited generalization abilities of their algorithm and the limited flexibility of their system. For example, the platform for OpenTeleVision~\cite{cheng2024opentv} and HATO~\cite{lin2024hato} does not support the movable base and waist, limiting the working space of the robot. HumanPlus~\cite{fu2024humanplus} and OmniH2O~\cite{he2024omnih2o} can whole-body teleoperate the humanoid robot, while the manipulation skills learned from their system are only limited to the training scene and can not generalize to other scenes due to the hardness in collect diverse data.  Maniwhere~\cite{yuan2024maniwhere} achieves real-world scene generalization on simple tabletop pushing tasks, while it is hard to apply their sim-to-real pipeline to humanoid robots due to the system complexity of humanoid robots. Similarly, 3D Diffusion Policy (DP3, ~\cite{Ze2024DP3}) only shows the object/view generalization with tabletop robot arms. The Robot Utility Model~\cite{etukuru2024robot_utility_model} also generalizes skills to the new environment with imitation learning, while they have to use data collected from 20 scenes for scene generalization, compared to only 1 scene we use.

% Image-based imitation learning methods, such as Diffusion Policy~\cite{chi2023diffusion_policy}, have achieved significant success~\cite{lin2024hato,ding2024bunnyvisionpro,yang2024equibot,wang2024equivariant,Ze2024DP3}, while their limited generalization abilities restrict their application in complex real-world environments. Several recent works aim to address these limitations~\cite{Ze2024DP3,yang2024equibot,wang2024gendp,wang2024equivariant,tian2024view}. Among these, the 3D Diffusion Policy (DP3, \cite{Ze2024DP3}) has demonstrated notable generalization abilities and broad applicability to diverse robotic tasks~\cite{he2024learning,ding2024bunnyvisionpro,yang2024equibot,yang2024ace}. Nonetheless, 3D visuomotor policies are inherently dependent on precise camera calibration and fine-grained point cloud segmentation~\cite{Ze2023GNFactor,Ze2024DP3,3d_diffuser_actor,qin2023dexpoint,tian2024view}, which limits their deployment on mobile platforms such as humanoid robots.  This work tackles this important problem and extends the application of 3D visuomotor policies into a more general setting.

In this paper, we take a significant step forward by building a real-world humanoid robot learning system that enables a full-sized humanoid robot to perform manipulation tasks in unseen real-world scenes, utilizing 3D visuomotor policies.

\section{Generalizable Humanoid Manipulation with 3D Diffusion Policies}
\label{sec: our system}

\begin{figure*}[htbp]
    \centering
\includegraphics[width=1.0\linewidth]{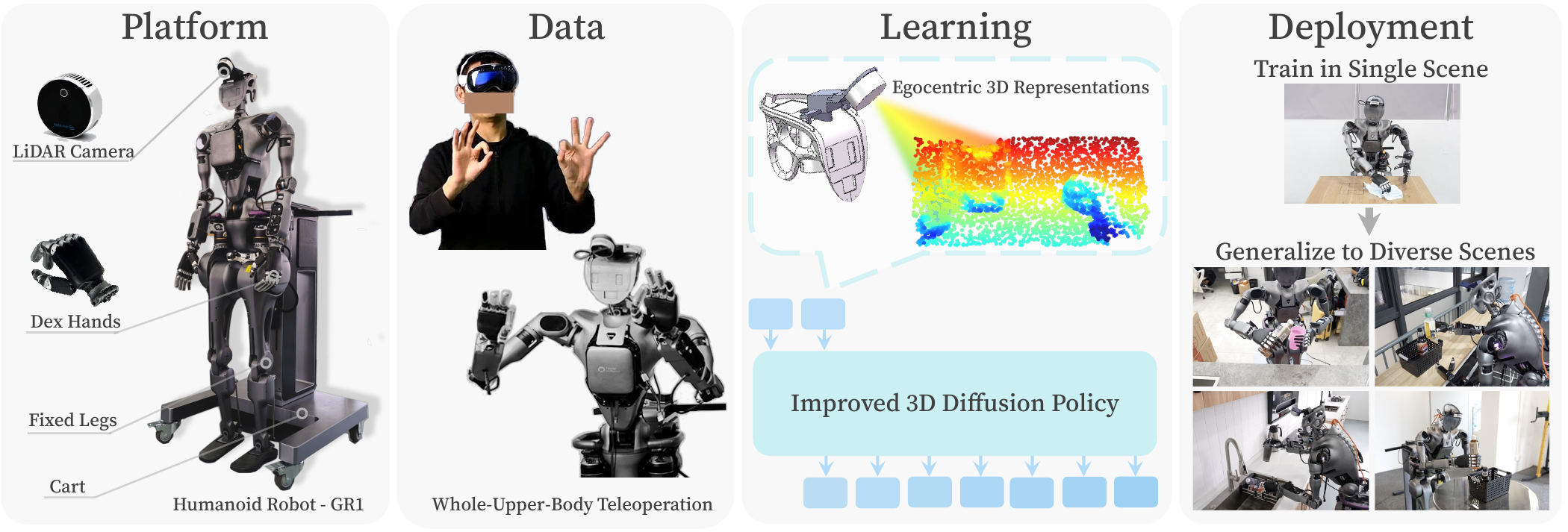}
 \vspace{-0.2in}
    \caption{\textbf{Overview of our system.} Our system mainly consists of four parts: the humanoid robot platform, the data collection system, the visuomotor policy learning method, and the real-world deployment. With this system, our humanoid robot performs autonomous skills in diverse real-world scenes.
    % For the learning part, we develop Improved 3D Diffusion Policy (iDP3) as a visuomotor policy for general-purpose robots. 
    }
    \label{fig:system}
    \vspace{-0.2in}
\end{figure*}

In this section, we present our real-world imitation learning system deployed on a full-sized humanoid robot. An overview of the system is provided in Figure~\ref{fig:system}. 

\subsection{Humanoid Robot Platform}
\label{sec: platform}

\noindent\textbf{Humanoid Robot.} We use Fourier GR1~\cite{Fourier_GR1}, a full-sized humanoid robot, equipped with two Inspire Hands~\cite{Inspire_Hand}. We enable the whole upper body \{\textit{head, waist, arms, hands}\}, totaling $25$ degrees-of-freedom (DoF). We disable the lower body for stability and instead use a cart for movement. Though previous systems such as HumanPlus~\cite{fu2024humanplus} and OmniH2O~\cite{he2024omnih2o} have shown the usage of humanoid legs, the loco-manipulation skills of these systems are still limited due to the hardware constraints. We emphasize that our system with 3D learning algorithms is general and could generalize to other humanoid robots with and without legs.

\noindent\textbf{LiDAR Camera.} To capture high-quality 3D point clouds, we utilize the RealSense L515~\cite{L515}, a solid-state LiDAR camera. The camera is mounted on the robot head to provide egocentric vision. Previous studies have demonstrated that cameras with less accurate depth sensing, such as the RealSense D435~\cite{D435}, can result in suboptimal performance for DP3~\cite{Ze2024DP3, wang2024rise}. It is important to note that, however, even the RealSense L515 does not produce perfectly accurate point clouds. We also try other LiDAR cameras such as Livox Mid-360, but we found that the resolution and the frequency of such LiDARs do not support contact-rich and real-time robotic manipulation.

\noindent\textbf{Height-Adjustable Cart.} A major challenge in generalizing manipulation skills to real-world environments is the wide variation in scene conditions, particularly \textit{the differing heights of tabletops}. To address this, we utilize a height-adjustable cart, eliminating the need for complex whole-body control. While this simplifies the manipulation process, we believe our approach will perform equally well once whole-body control techniques become more mature.

\subsection{Human-Like Robot Data}
\label{sec: data}

\noindent\textbf{Whole-Upper-Body Teleoperation.} To obtain human-like humanoid robot data, we design a teleoperation system that can teleoperate the robot's entire upper body, including the head, waist, hands, and arms. We use the Apple Vision Pro (AVP, \cite{apple_vision_pro}) to obtain accurate and real-time human data, \textit{e.g.}, the 3D positions and orientations of the head/hands/wrists~\cite{park2024avp}. With this human data, we compute the corresponding robot joint angles respectively. More specifically, 1) the robot arm joints are computed with Relaxed IK~\cite{rakita2018relaxedik} to track human wrist positions; 2) the robot waist and head joints are computed by using the rotation of the human head. We also stream the real-time robot vision back to humans for immersive teleoperation feedback~\cite{cheng2024opentv}.

\noindent\textbf{Latency of Teleoperation.} The use of a LiDAR sensor significantly occupies the bandwidth/CPU of the onboard computer, resulting in a teleoperation latency of approximately 0.5 seconds. We also try two LiDAR sensors (one additionally mounted on the wrist), which introduce extremely high latency and thus make the data collection infeasible.

\noindent\textbf{Data for Learning.} We collect trajectories of observation-action pairs during teleoperation, where observations consist of two parts: 1) visual data, such as point clouds and images, and 2) proprioceptive data, such as robot joint positions. Actions are represented by the target joint positions. We also tried using end-effector poses as proprioceptions/actions, finding that directly applying joint positions as action space is more accurate, mainly due to the noise in the real world to compute the end-effector poses.

\subsection{Improved 3D Diffusion Policy}
\label{sec: iDP3}
\noindent\textbf{3D Diffusion Policy (DP3, \cite{Ze2024DP3})} is an effective 3D visuomotor policy that marries sparse point cloud representations with diffusion policies. Although DP3 has shown impressive results across a wide range of manipulation tasks, it is not directly deployable on general-purpose robots such as humanoid robots or mobile manipulators due to its inherent dependency on precise
camera calibration and fine-grained point cloud segmentation. Furthermore, the accuracy of DP3 requires further improvements for effective performance in more complex tasks. In the following, we detail several modifications to achieve targeted improvements. The resulting improved algorithm is termed as the \textit{Improved 3D Diffusion Policy (iDP3)}.

\begin{figure}[t!]
    \centering
\includegraphics[width=0.7\linewidth]{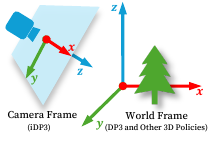}
    \vspace{-0.2in}
\caption{\textbf{iDP3 utilizes 3D representations in the camera frame,} while the 3D representations of other recent 3D policies including DP3~\cite{Ze2024DP3}  are in the world frame, which relies on accurate camera calibration and can not be extended to mobile robots.
    }
    \label{fig:camera frame}
    \vspace{-0.2in}
\end{figure}

\noindent\textbf{Egocentric 3D Visual Representations.} DP3 leverages a 3D visual representation in the world frame, enabling easy segmentation of the target object~\cite{Ze2024DP3,wang2024dexcap}. However, for general-purpose robots like humanoids, the camera mount is not fixed, making camera calibration and point cloud segmentation impractical. To tackle this problem, we propose directly using the 3D representation from the camera frame, as shown in Figure~\ref{fig:camera frame}. We term this class of 3D representations as \textit{egocentric 3D visual representations}.

\noindent\textbf{Scaling Up Vision Input.} Leveraging egocentric 3D visual representations presents challenges in eliminating extraneous point clouds, such as backgrounds or tabletops, especially without relying on foundation models.
To mitigate this, we propose a straightforward but effective solution: scaling up the vision input. Instead of using standard sparse point sampling as in previous systems~\cite{Ze2024DP3, wang2024dexcap, yang2024equibot}, we significantly increase the number of sample points to capture the entire scene. Despite its simplicity, this approach proves to be effective in our real-world experiments.

\noindent\textbf{Improved Visual Encoder.} We replace the MLP visual encoder in DP3 with a pyramid convolutional encoder. We find that convolutional layers produce smoother behaviors than fully-connected layers when learning from human data, and incorporating pyramid features from different layers further enhances accuracy.

\noindent\textbf{Longer Prediction Horizon.} The jittering from human experts and the noisy sensors exhibit much difficulty in learning from human demonstrations, which causes DP3 to struggle with short-horizon predictions. By extending the prediction horizon, we effectively mitigate this issue.

\noindent\textbf{Implementation Details.} For the optimization,  we train 300 epochs for iDP3 and all other methods with AdamW~\cite{loshchilov2018adamw}. For the diffusion process, we use 50 training steps and 10 inference steps with DDIM~\cite{song2020ddim}. For the 
point cloud sampling, we replace farthest point sampling (FPS) used in DP3~\cite{Ze2024DP3} with 
a cascade of voxel sampling and uniform sampling, which ensures  the sampled points cover the 3D space with a faster inference speed.

\subsection{Real-World Deployment}
We train iDP3 on our collected human demonstrations. Notably, we do not rely on camera calibration or manual point cloud segmentation as mentioned before. Therefore, our iDP3 policy can be seamlessly transferred to new scenes without requiring additional efforts such as calibration/segmentation. Besides, iDP3 performs real-time inference (15hz) with only onboard robot computing, making the deployment to the open world accessible.

\begin{figure*}[t]
    \centering
\includegraphics[width=1.0\linewidth]{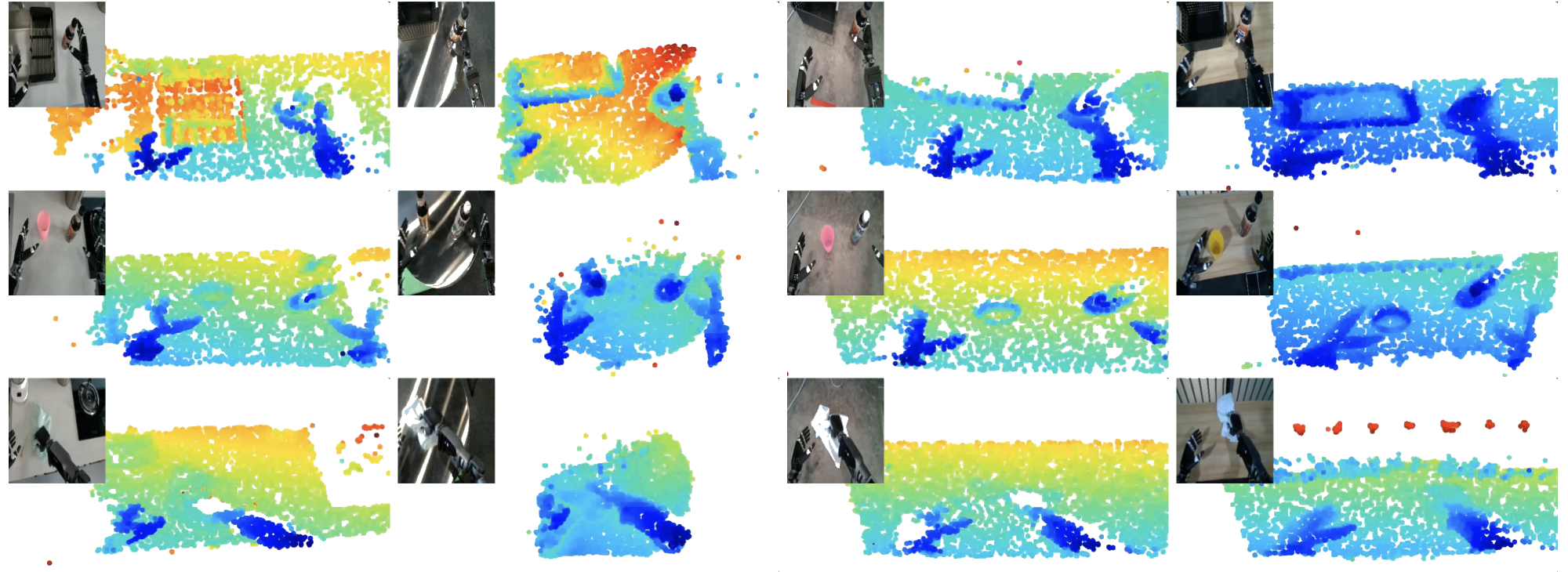}
\vspace{-0.3in}
    \caption{\textbf{Visualization of egocentric 2D and 3D observations.} This figure highlights the complexity of diverse real-world scenes. Videos are available on \website.}
    \label{fig:2d and 3d obs vis}
    \vspace{-0.2in}
\end{figure*}

\section{Experiments and Analysis}

To evaluate the effectiveness of our system, we conduct extensive real-world ablations with our system. We select the \textit{Pick\&Place} task as the primary benchmark for our analysis, and further showcase the \textit{Pick\&Place}, \textit{Pour}, and \textit{Wipe} tasks in diverse unseen scenarios.

\subsection{Experiment Setup}
\noindent\textbf{Task Description.} In this task, the robot grasps a lightweight cup and moves it aside. The challenge for humanoid robots with dexterous hands is that the cup is similar in size to the hands; thus, even small errors result in collisions or missed grasps. This task requires more precision than using parallel grippers, which can open wider to avoid collisions.

\noindent\textbf{Task Setting.} We train the Pick\&Place task under four settings: \{1st-1, 1st-2, 3rd-1, 3rd-2\}.  ``1st" uses an egocentric view, and ``3rd" uses a third-person view. The numbers behind represent the number of demonstrations used for training, with each demonstration consisting of 20 rounds of successful execution. The training dataset is kept small to highlight the differences between methods. The object position is randomly sampled in a 10cm$\times$20cm region.

\noindent\textbf{Evaluation Metric.} We run three episodes for each method, each consisting of 1,000 action steps. In total, each method is evaluated with around $130$ trials, ensuring a thorough evaluation of each method.
We record both the number of successful grasps and the total number of grasp attempts. The successful grasp count reflects the accuracy of the policy. The total number of attempts serves as a measure of the policy's smoothness, since the jittering policies tend to hang around and have few attempts as we observe in experiments.

\begin{figure}[t]
    \centering
\includegraphics[width=1.0\linewidth]{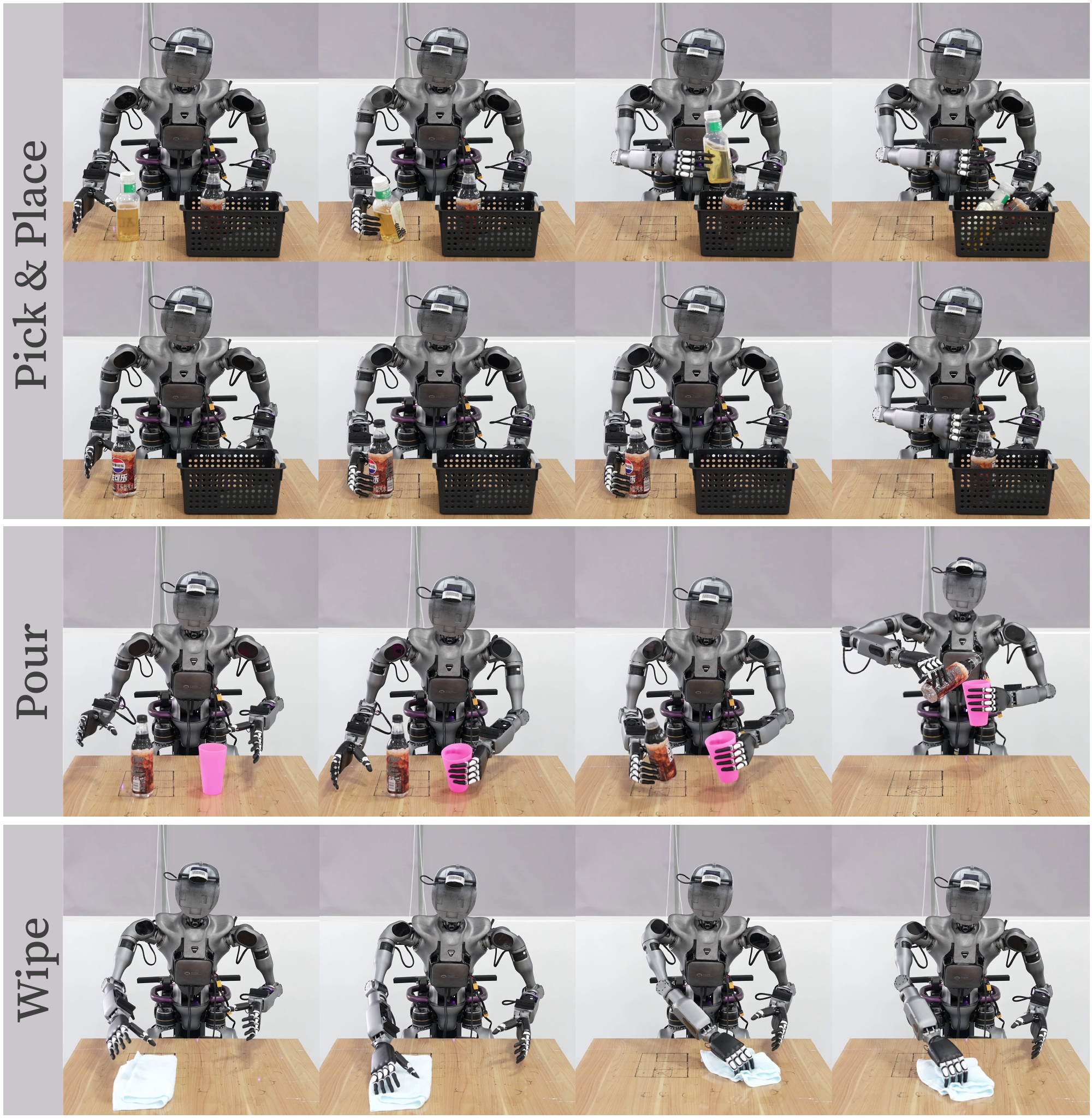}
    \vspace{-0.2in}
    \caption{\textbf{Trajectories of our three tasks in the training scene,} including Pick\&Place, Pour, and Wipe. We carefully select daily tasks so that the objects are common in daily scenes and the skills are useful across scenes.}
    \label{fig:task vis}
    \vspace{-0.2in}
\end{figure}

\begin{figure*}[t!]
    \centering
\includegraphics[width=1.0\linewidth]{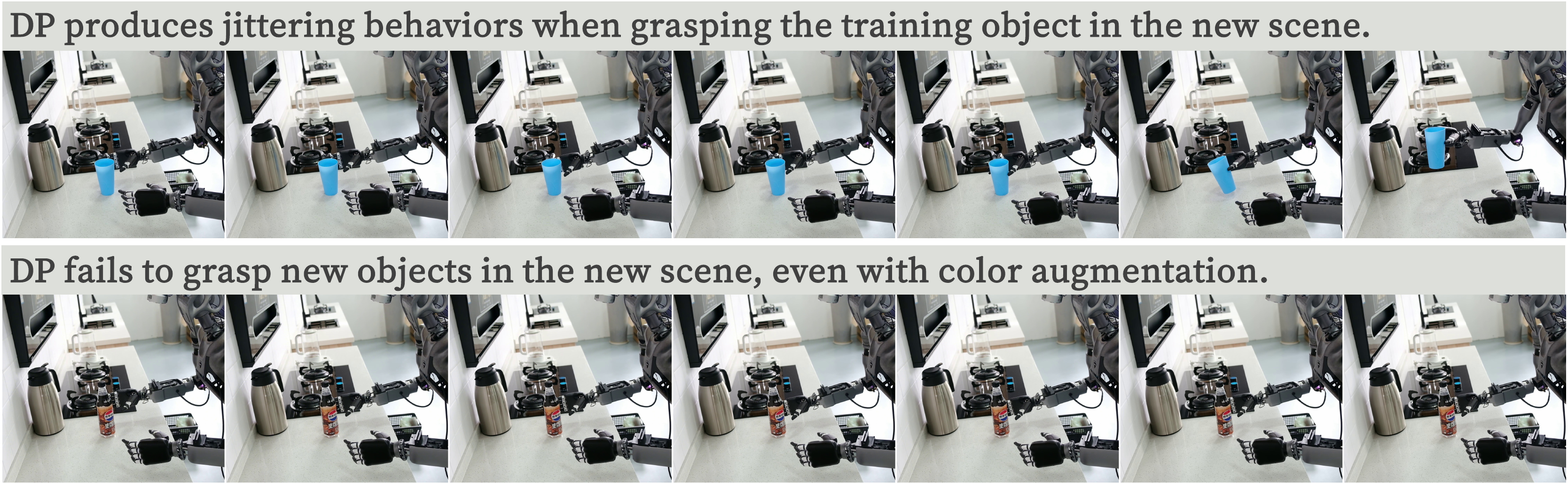}
    \vspace{-0.2in}
    \caption{\textbf{Failure cases of image-based methods in new scenes.} Here DP corresponds to \textbf{DP (\ding{86}R3M)} in Table~\ref{table: compare to baselines}, which is the strongest image-based baseline we have. We find that even added with color augmentation during training, image-based methods still struggle in the new scene/object.}
    \label{fig:baseline vis}
    \vspace{-0.1in}
\end{figure*}

\begin{figure}[htbp]
    \centering
\includegraphics[width=0.8\linewidth]{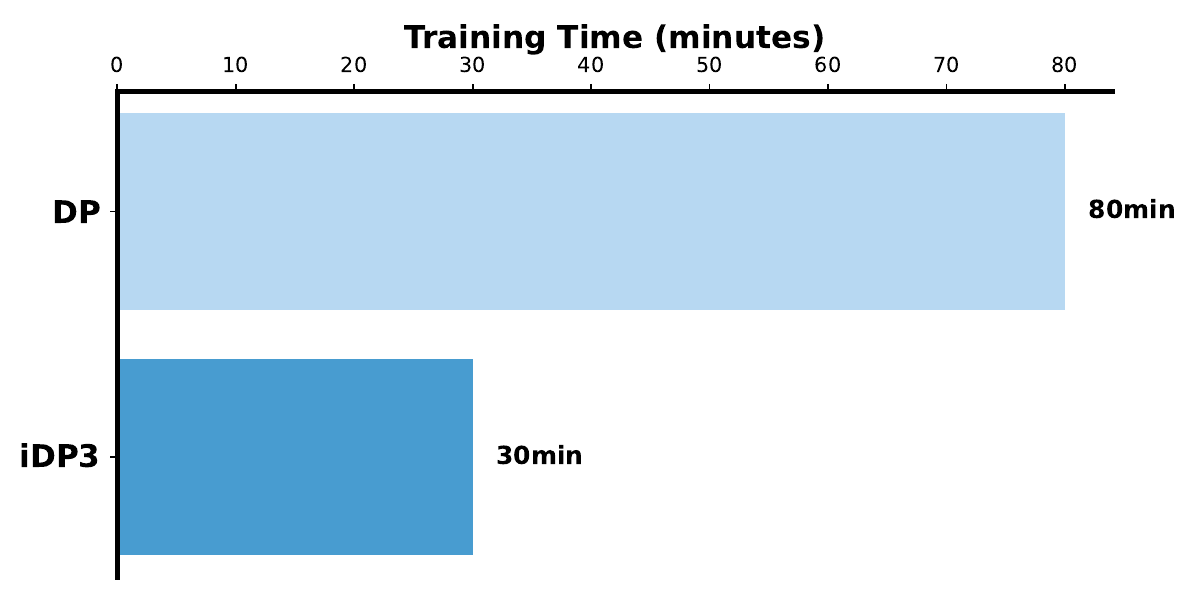}
\vspace{-0.1in}
\caption{\textbf{Training time.} Due to using 3D representations, iDP3 saves training time compared to Diffusion Policy (DP), even after we scale up the 3D vision input. This advantage becomes more evident when the number of demonstrations gets large.}
    \label{fig:training time}
    \vspace{-0.1in}
\end{figure}

\subsection{Effectiveness}

\begin{table}[t]
\centering
%\vspace{-0.1in}
\caption{\textbf{Efficiency of \ours compared to baselines.}  To improve the robustness of the baselines, we have added Random Crop and Color Jitter augmentation to all image-based methods during training. \textbf{All the methods are evaluated with more than 100 trials,} ensuring less randomness in real-world evaluation. Without modification, original DP~\cite{chi2023diffusion_policy} and DP3~\cite{Ze2024DP3} work badly on our humanoid robot.}
%%JP: is augmentation during training or at test time?
% ze: sure! during training.
\label{table: compare to baselines}
\vspace{-0.05in}

\resizebox{0.5\textwidth}{!}{%
\begin{tabular}{l|cccccccccc}
\toprule[1.5pt]

 \multirow{2}{*}{\textbf{Baselines}}   & \multirow{2}{*}{DP} & \multirow{2}{*}{DP3} & DP &  DP& iDP3 & \multirow{2}{*}{\textbf{\ours}} \\
 &  & & (\ding{100}R3M) & (\ding{86}R3M) &  (DP3 Encoder)\\
\midrule
1st-1 & 0/0 & 0/0 &  11/33 & 24/39 & 15/34 & 21/38\\
1st-2  & 7/34 & 0/0 & 10/28 & 27/36 &12/27 & 19/30\\
3rd-1  & 7/36 & 0/0 & 18/38 & 26/38 &15/32 & 19/34\\
3rd-2 & 10/36 & 0/0 &  23/39 & 22/34 &16/34 & 16/37 \\ 
\midrule
Total  & 24/106 & 0/0 &  62/138 & \textbf{99/147} & 58/127 & \textbf{75/139} \\
\bottomrule
\end{tabular}}
\vspace{-0.2in}
\end{table}

%%JP: lets keep all figures and tables at the top of a page. It's a bit messy to embed a table in the middle of the text

We compare iDP3 with several strong baselines, including: a) \textbf{DP}: Diffusion Policy~\cite{chi2023diffusion_policy} with a ResNet18 encoder; b) \textbf{DP (\ding{100}R3M)}: Diffusion Policy with a frozen R3M~\cite{nair2022r3m} encoder; c) \textbf{DP (\ding{86}R3M)}: Diffusion Policy with a finetuned R3M encoder; d) original DP3 without any modifications; and e) \textbf{iDP3 (DP3 Encoder)}: iDP3 using the DP3 encoder~\cite{chi2023diffusion_policy}. All image-based methods use the same policy backbone as iDP3 and Random Crop and
Color Jitter augmentations to improve robustness and generalization. The RGB image resolution is $224\times224$, resized from the raw image from the RealSense camera.

The results, presented in Table~\ref{table: compare to baselines}, show that iDP3 significantly outperforms vanilla DP and DP3, DP with a frozen R3M encoder, and iDP3 with the DP3 encoder. However, we find that DP with a finetuned R3M is a particularly strong baseline, outperforming iDP3 in these settings. We hypothesize that this is because finetuning pre-trained models are often more effective compared to training-from-scratch ~\cite{hansen2022lfs}, and there are currently no similar pre-trained 3D visual models for robotics.

Though DP+finetuned R3M is more effective in these settings, we find that image-based methods are overfitting to the specific scenario and object, failing to generalize to wild scenarios, as shown in Section~\ref{sec: capbilities}.

Additionally, we believe there is still room for improvement in iDP3. Our current 3D visual observations are quite noisy due to the limitations of the sensing hardware. We expect that more accurate 3D observations could lead to optimal performance in 3D visuomotor policies, as demonstrated in simulation~\cite{Ze2024DP3}.

\begin{table}[t]
%\vspace{-0.10in}
\centering
\caption{\textbf{Ablation on \ours.} The results demonstrate that removing certain key modifications from iDP3 significantly impacts the performance of DP3, leading to either failure in learning from human data or reduced accuracy. \textbf{All the methods are evaluated with more than 100 trials,} ensuring less randomness in real-world evaluation.}
\label{table: all ablations}
\vspace{-0.05in}

\resizebox{0.5\textwidth}{!}{%
\begin{tabular}{l|cccc|ccccc}
\toprule[1.5pt]

  \textbf{Visual Encoder}  & 1st-1 & 1st-2 & 3rd-1 & 3rd-2  & Total\\

  \midrule
Linear (DP3) & 15/34 & 12/27 & 15/32 & 16/34 & 58/127 \\
Conv& 9/33 & 14/32 & 14/33 & 12/33 & 49/131\\
Linear+Pyramid & 15/34 & \textbf{20/31} & 13/33 & \textbf{18/36} & 66/134 \\
\textbf{Conv+Pyramid (iDP3)} & \textbf{21/38} & 19/30 & \textbf{19/34} & 16/37 & \textbf{75/139}\\
\bottomrule
\end{tabular}}

\vspace{0.05in}

\resizebox{0.5\textwidth}{!}{%
\begin{tabular}{l|cccc|ccccc}
\toprule[1.5pt]

  \textbf{Number of Points}  & 1st-1 & 1st-2 & 3rd-1 & 3rd-2 & Total\\

  \midrule

1024 (DP3) & 11/28 & 10/30 & 18/35 & 17/36 & 56/129\\
2048 & 17/35 & 13/28 & 17/32 & \textbf{18/33} & 65/128\\
\textbf{4096 (iDP3)}  & 21/38 & \textbf{19/30} & \textbf{19/34 }& 16/37 & \textbf{75/139}\\
8192 & \textbf{24/35} & 16/28 & 14/33 & \textbf{18/36} & 72/132\\
\bottomrule
\end{tabular}}

\vspace{0.05in}

\resizebox{0.5\textwidth}{!}{%
\begin{tabular}{l|cccc|ccccc}
\toprule[1.5pt]

 \textbf{Prediction Horizon}  & 1st-1 & 1st-2 & 3rd-1 & 3rd-2 & Total \\

  \midrule

4 (DP3) & 0/0 & 0/0 & 0/0 & 0/0 & 0/0 \\
8 & 0/0 & 3/18 & 18/36 & 12/34 & 33/88\\
\textbf{16 (iDP3)}  & \textbf{21/38} & 19/30 & \textbf{19/34} & \textbf{16/37} & \textbf{75/139}\\
32 & 9/34 & \textbf{20/30} & 14/33 & 12/33 & 55/130 \\
\bottomrule
\end{tabular}}

\vspace{-0.22in}
\end{table}

\begin{table*}[ht!]
\centering
\caption{\textbf{Capabilities of iDP3.} While iDP3 maintains similar efficiency to DP (\ding{86}R3M) (abbreviated as DP), it stands out with remarkable generalization capabilities, making it well-suited for real-world deployment. For evaluation in the new scene, we use the kitchen scene shown in Figure~\ref{fig:baseline vis} and unseen objects are also included. We do not test Wipe in generalization settings since Wipe is achieved with high success rates for all methods. We do not conduct more evaluation on baselines in other unseen real-world scenes as we find the baselines can not work in unseen scenes, same as what we observe in the kitchen scene.}
\label{table: capability}
\vspace{-0.05in}

\resizebox{1.0\textwidth}{!}{%
\begin{tabular}{l|ccccccccc}
\toprule[1.5pt]

 \textbf{Training}  & DP & \textbf{\ours} & \\

  \midrule
Pick\&Place & \textbf{9/10} & \textbf{9/10} & \\
Pour & \textbf{9/10} & \textbf{9/10} & \\
Wipe & \textbf{10/10} & \textbf{10/10} \\
\bottomrule
\end{tabular}
\begin{tabular}{l|ccccccccc}
\toprule[1.5pt]

 \textbf{New Object} & DP & \textbf{\ours} \\

  \midrule
  Pick\&Place & 3/10 & \textbf{9/10}\\

Pour & 1/10 & \textbf{9/10}  \\
Wipe & -- & -- \\
\bottomrule
\end{tabular}
\begin{tabular}{l|ccccccccc}
\toprule[1.5pt]

 \textbf{New View} & DP & \textbf{\ours} \\

  \midrule
  Pick\&Place & 2/10 & \textbf{9/10}\\
Pour & 0/10 & \textbf{9/10}  \\
Wipe & -- & -- \\
\bottomrule
\end{tabular}
\begin{tabular}{l|ccccccccc}
\toprule[1.5pt]

 \textbf{New Scene} & DP & \textbf{\ours} \\

  \midrule
  Pick\&Place & 2/10 & \textbf{9/10}\\
Pour & 1/10 & \textbf{9/10}  \\
Wipe & -- & -- \\
\bottomrule
\end{tabular}
}
\vspace{-0.1in}
\end{table*}

\subsection{Ablations}
We conduct ablation studies on several modifications to DP3, including improved visual encoders, scaled visual input, and a longer prediction horizon. Our results, given in Table~\ref{table: all ablations}, demonstrate that without these modifications DP3 either fails to learn effectively from human data or exhibits significantly reduced accuracy. 

More specifically, we observe that 1) our improved visual encoder could both improve the smoothness and accuracy of the policy; 2) scaled vision inputs are helpful, while the performance gets saturated in our tasks with more points; 3) an appropriate prediction horizon is critical, without which DP3 fails to learn from human demonstrations. 

Additionally, Figure~\ref{fig:training time} presents the training time for iDP3, demonstrating a significant reduction compared to Diffusion Policy. This efficiency is maintained even when the number of point clouds increases to several times that of DP3~\cite{Ze2024DP3}.

\subsection{Capabilities}
\label{sec: capbilities}

In this section, we show more generalization capabilities of our system on humanoid robots. We also conduct more comparisons between iDP3 and DP (\ding{86}R3M) (abbreviated as DP in this section) and show that iDP3 is more applicable in the challenging and complex real world. Results are given in Table~\ref{table: capability}.

\noindent\textbf{Tasks.} We select three tasks, \textit{Pick\&Place}, \textit{Pour}, and \textit{Wipe}, to demonstrate the capabilities of our system. We ensure that these tasks are common in daily life and could be useful for humans.
For instance, Pour is frequently performed in restaurants, and Wipe in cleaning tables in households.

\noindent\textbf{Data.} For each task, we collect 10 demonstrations$\times$10 rollouts, totalling 300 episodes for all tasks. For Pick\&Place and Pour, the object poses are randomized in a region of 10cm$\times$10cm. 

% We do not collect data in a larger region, since we find that a larger task region simply requires more data~\cite{zhao2024aloha}. 

% Besides, collecting large-scale data is not feasible due to the usage of AVP.

\noindent\textbf{Effectiveness.} As shown in Table~\ref{table: capability}, both iDP3 and DP achieve high success rates in the training environment with the training objects.

\begin{figure}[t!]
    \centering
\includegraphics[width=1.0\linewidth]{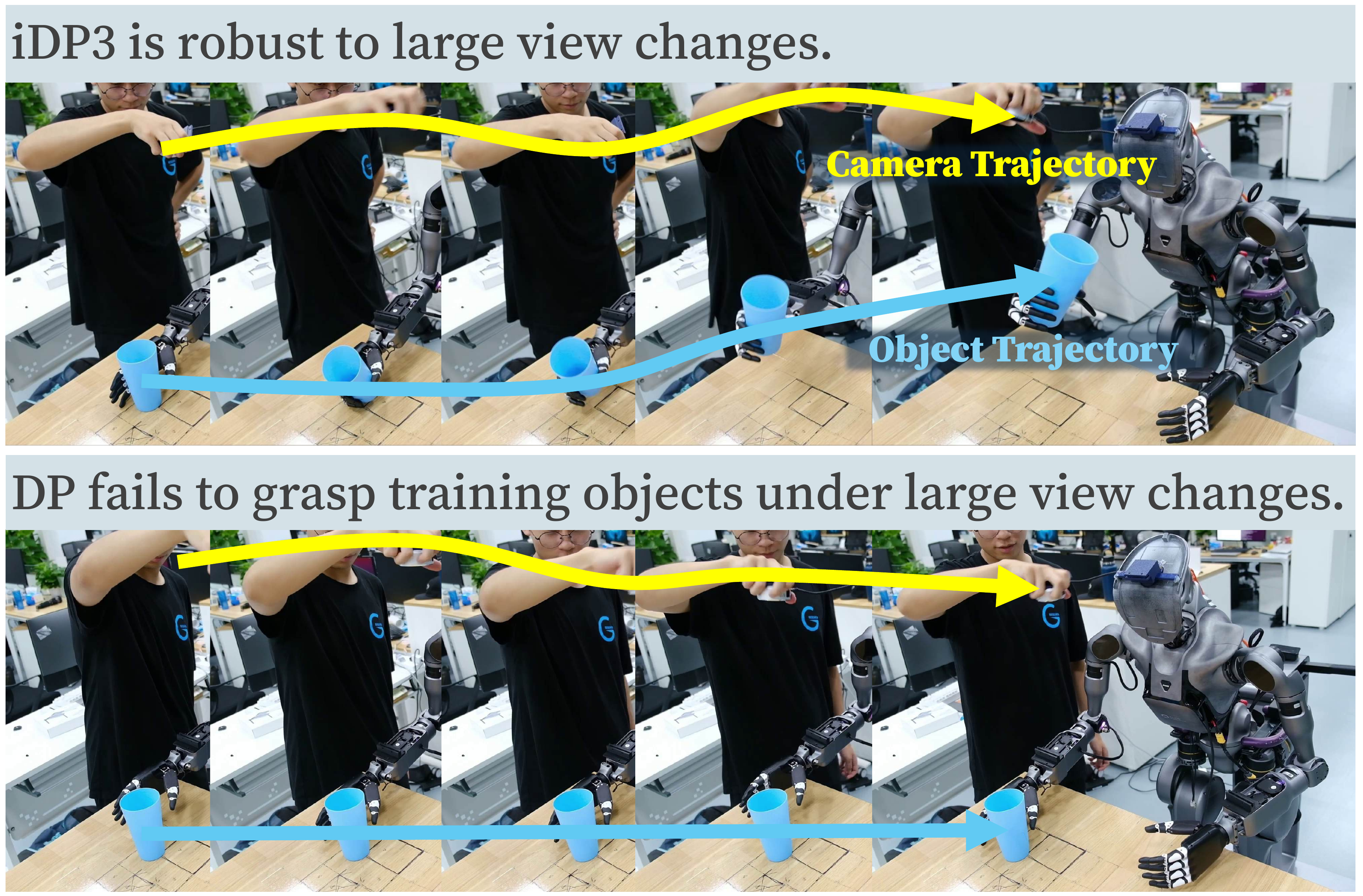}
\vspace{-0.2in}
    \caption{\textbf{View invariance of iDP3.} We find that egocentric 3D representations are surprisingly view-invariant.
    Here DP corresponds to \textbf{DP (\ding{86}R3M)} in Table~\ref{table: compare to baselines}, which is the strongest image-based baseline we have.}
    \label{fig:view invariance}
    \vspace{-0.2in}
\end{figure}

\noindent\textbf{Property 1: View Invariance.} Our egocentric 3D representations demonstrate impressive view invariance. As shown in Figure~\ref{fig:view invariance}, iDP3 consistently grasps objects even under large view changes, while DP struggles to grasp even the training objects. DP shows occasional success only with minor view changes. Notably, unlike recent works \cite{tian2024view,yang2024equibot,wang2024equivariant}, we did not incorporate specific designs for equivariance or invariance.

\begin{figure}[htbp!]
    \centering
\includegraphics[width=0.5\linewidth]{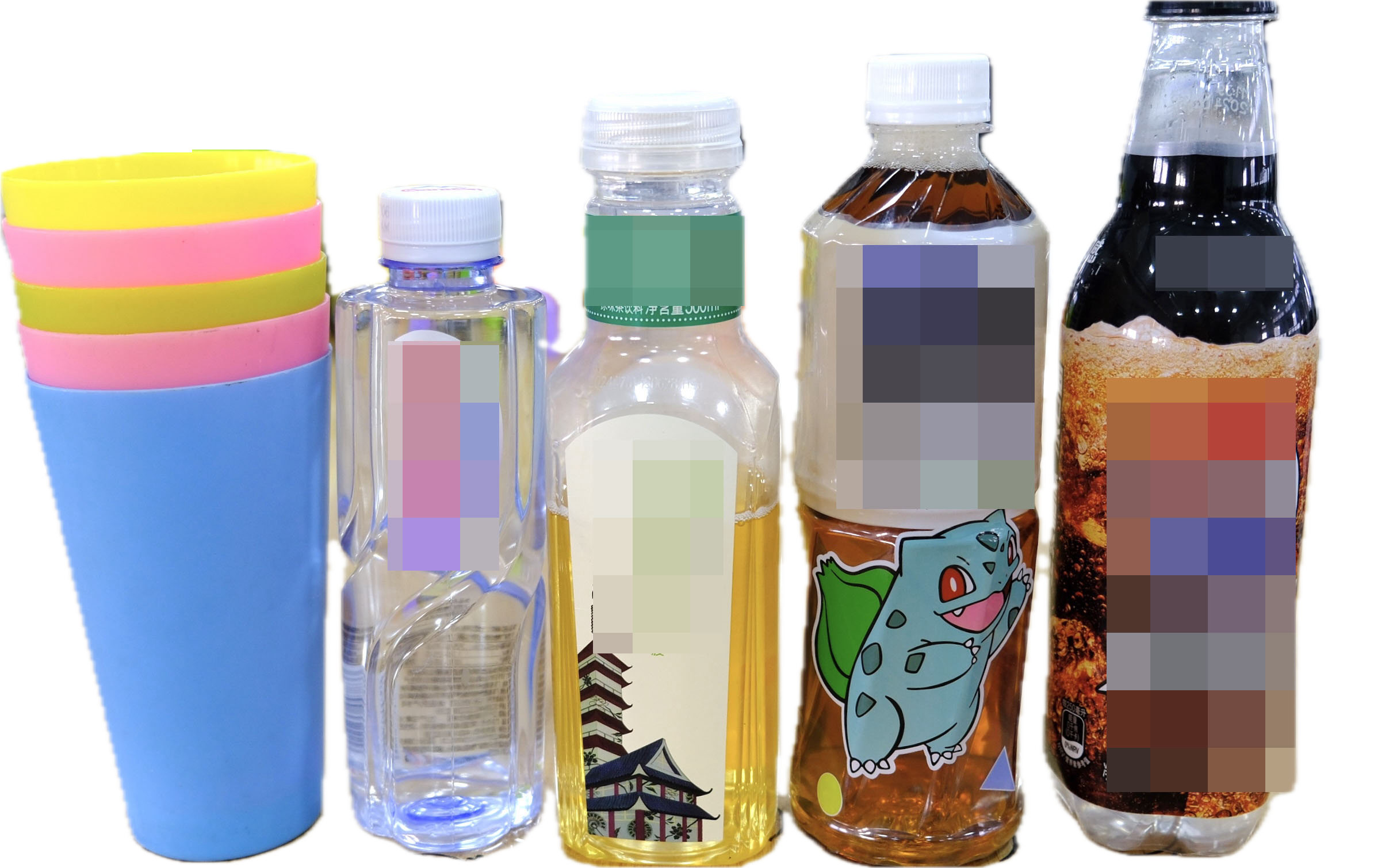}
    \caption{\textbf{Objects used in Pick\&Place and Pour.} We only use the cups as the training objects, while our method naturally handles other unseen bottles/cups.}
    \label{fig:objects}
    \vspace{-0.15in}
\end{figure}

\noindent\textbf{Property 2: Object Generalization.} We evaluated new kinds of cups/bottles beside the training cup, as shown in Figure~\ref{fig:objects}. 
While DP, due to the use of Color Jitter augmentation, can occasionally handle unseen objects, it does so with a low success rate. In contrast, iDP3 naturally handles a wide range of objects, thanks to its use of 3D representations.

\noindent\textbf{Property 3: Scene Generalization.} We further deploy our policy in various real-world scenarios, as shown in Figure~\ref{fig:diverse scene}. These scenes are nearby the lab and \textit{none of the scenes are cherry-picked.} The real world is far noisier and more complex than the controlled tabletop environments used in the lab, leading to reduced accuracy for image-based methods (Figure~\ref{fig:baseline vis}). Unlike DP, iDP3 demonstrates surprising robustness across all scenes. Additionally, we provide visualizations of both 2D and 3D observations in Figure~\ref{fig:2d and 3d obs vis}.

\section{Conclusions and Limitations}

\noindent\textbf{Conclusions.} This work presents a real-world imitation learning system that enables a full-sized humanoid robot to generalize practical manipulation skills to diverse real-world environments, trained with data collected solely in one single scene. With more than 2000 rigorous evaluation trials, we present an improved 3D Diffusion Policy, that can learn robustly from human data and perform effectively on our humanoid robot. The results that our humanoid robot can perform autonomous manipulation skills in diverse real-world scenes show the potential of using 3D visuomotor policies in real-world manipulation tasks with data efficiency.

\noindent\textbf{Limitations.} 1) Teleoperation with Apple Vision Pro is easy to set up, but it is tiring for human teleoperators, making imitation data hard to scale up within the research lab. 2) The depth sensor still produces noisy and inaccurate point clouds, limiting the performance of iDP3. 3) Collecting fine-grained manipulation skills, such as turning a screw, is time-consuming due to teleoperation with AVP; systems like Aloha~\cite{zhao2023aloha} are easier to collect dexterous manipulation tasks at this stage.
4) We avoided using the robot's lower body, as maintaining balance is still challenging due to the hardware constraints brought by current humanoid robots. In general, scaling up high-quality manipulation data is the main bottleneck. In the future, we hope to explore how to scale up the training of 3D visuomotor policies with more high-quality data and how to employ our 3D visuomotor policy learning pipeline to humanoid robots with whole-body control.

\section*{Acknowledgments}
We would like to thank Jie Gu, Bin Zhou, and Yusheng Cai from Fourier Intelligence for hardware support, Yuxiang Gao for help in teleoperation, Shuo Hu for help in the 3D printing of the camera mount, Zeqian Bao, Renzhi Tao, and Jiayan Gu for helpful discussions. Besides, we would like to thank Chen Wang, Yunzhi Zhang, Zizhang Li, and Haoyu Xiong from Stanford University for their insightful discussions. This work is in part supported by ONR MURI N00014-22-1-2740, ONR MURI N00014-24-1-2748, and the Okawa Foundation.
\bibliographystyle{IEEEtran}
\bibliography{main}

\end{document}